\documentclass[11pt]{article}
\usepackage{graphicx}
\usepackage{multirow}
\usepackage{EACL2023}
    
\usepackage{graphicx}
\usepackage{multirow}

\usepackage{times}
\usepackage{latexsym}
\usepackage{enumitem}
\usepackage{booktabs}
\usepackage{tcolorbox}
\usepackage{flushend}

\usepackage[T1]{fontenc}

\usepackage[utf8]{inputenc}

\usepackage{microtype}

\usepackage{inconsolata}

\begin{document}

\title{MMT: A Multilingual and Multi-Topic Indian Social Media Dataset}

\author{Dwip Dalal \and ... \and Mayank Singh \\
        Address line \\ ... \\ Address line}

\author{Dwip Dalal$^{1}$, Vivek Srivastava$^{2}$, Mayank Singh$^{1}$ \\
$^{1}$IIT Gandhinagar, Gandhinagar, India \\
$^{2}$TCS Research, Pune, India \\
\texttt{\{dwip.dalal, singh.mayank\}@iitgn.ac.in} \\
\texttt{srivastava.vivek2@tcs.com}
}

\maketitle
\begin{abstract}
Social media plays a significant role in cross-cultural communication. A vast amount of this occurs in code-mixed and multilingual form, posing a significant challenge to Natural Language Processing (NLP) tools for processing such information, like language identification, topic modeling, and named-entity recognition. To address this, we introduce a large-scale multilingual, and multi-topic dataset (\textit{MMT}) collected from Twitter ($\approx$1.7 million Tweets), encompassing 13 coarse-grained and 63 fine-grained topics in the Indian context. We further annotate a subset of 5,346 tweets from the \textit{MMT} dataset with various Indian languages and their code-mixed counterparts. Also, we demonstrate that the currently existing tools fail to capture the linguistic diversity in \textit{MMT} on two downstream tasks, i.e., \textit{topic modeling} and \textit{language identification}. To facilitate future research, we will make the anonymized and annotated dataset available in the public domain.
\end{abstract}

\section{Introduction}
\label{sec: intro}
In the last decade, we have observed high growth in the number of available social media platforms, as well as the user engagement on these platforms~\cite {liu2014tweets}. Such widespread usage of these platforms makes them the primary means of information spread within as well as across cultures in any socially engaging event such as elections \cite{jungherr2016twitter}, entertainment \cite{antelmi2018understanding}, sports \cite{wang2020building}, science \cite{lopez2018social}, and technology \cite{kreiss2018technology}.

India, with a population of over 1.3 billion, attracts the attention of all major social media firms\cite{aneez2019india}; various studies \cite{bharucha2018social, singh2019analyzing} reaffirm the active participation of Indians on these platforms. With diversity and multilingualism deeply ingrained in the culture of India \cite{ishwaran1969multilingualism}, it is no wonder that we find huge volumes of code-mixed data \cite{thara2018code} in the Indian social media space -- which consequently makes it a goldmine for the NLP research community\cite{conway2019recent}.

The NLP community has always been interested in solving problems in multilinguality \cite{xue2021mt5} and multi-topicality \cite{yuan2018multilingual}. In most of the research, the two problems are addressed separately. However, several interesting questions emerge in multilingual-multitopical datasets. Here, we explore three research questions:
\begin{itemize}[noitemsep,nolistsep,leftmargin=*]
    \item \textit{\textbf{RQ1}}: how traditional topic modeling tools perform in multilingual settings?
    \item \textit{\textbf{RQ2}}: can we achieve better topic modeling with the multilingual data using the contextual topic models?
    \item \textit{\textbf{RQ3}}: how do multilingual language identification tools perform in multi-topical text?
\end{itemize}


To the best of our knowledge, we have not found extensive investigation into the answers to the above questions. This paper explores these pertinent questions supported by robust evaluations and presents interesting anecdotal examples.  



\section{Constructing The Multilingual and Multi-topic  Dataset}
\label{sec:dataset}

\subsection{MMT} The large-scale multilingual and multi-topic dataset is constructed in four phases as listed below:

\begin{enumerate}[noitemsep,nolistsep,leftmargin=*]

        \item \textbf{Annotator selection and grouping}: We selected a diverse group of 49 students who were either undergraduates, masters, or postgraduates from different regions and cultural backgrounds in India. These students hailed from various states across India, representing different parts of the country from north to south, east to west. The 49 students were self-organized into 13 teams, with 10 teams consisting of 4 members each and 3 teams consisting of 3 members each. All the students were native Indians and active Twitter users with high proficiency in English and knowledge of at least one Indian language. This selection criterion ensured a diverse and representative sample.

\item \textbf{Topic identification}:  As an initial step, we identify 13 topics relevant to the Indian context to capture and cater to various dimensions of discussions on social media, specifically Twitter. We enlist all 13 topics in Table \ref{tab: topics_dist}. The choice of seed topics is also motivated by the most frequently discussed and relevant areas to the Indian community, as it helps get quality large-scale data easily from Twitter.

\item \textbf{Subtopic selection}: Next, we collect the fine-grained categorization for each of the 13 seed topics. We assign one seed topic to each team and ask them to develop a set of subtopics within each seed topic. The teams have the flexibility to do their own study (within and outside the Twitter community) to come up with a set of subtopics. We provided teams with constructive feedback and suggestions for improvement to ensure the accuracy and relevance of the selected subtopics. We fostered a collaborative process to arrive at a consensus on 63 subtopics that encompass diversity and exhaustiveness. The selected subtopics for all 13 seed topics are presented in  Table \ref{tab: topics_dist}.

\item \textbf{Data collection}:  We curate data from Twitter based on the assigned subtopics for each seed topic. For this task, we employ the same set of 13 teams with a task of scraping at least 100K tweets (and the associated data and metadata) using the TWINT tool\footnote{https://github.com/twintproject/twint}
. The teams are encouraged and rewarded to curate more than 100K tweets. We further preprocess and remove the tweets with missing values. 

In total, MMT comprises 1,755,145 tweets, with 135K tweets on average for each topic (Table \ref{tab: topics_dist}). We  observe a high degree of multilingualism, with tweets coming from 47 languages (as identified by Twitter). Based on manual inspection, we observe that the Twitter language identification system (hereafter \textit{``TLID''}) assigns incorrect language tags to a large number of non-English tweets. 
\end{enumerate}

\begin{figure}[!t]
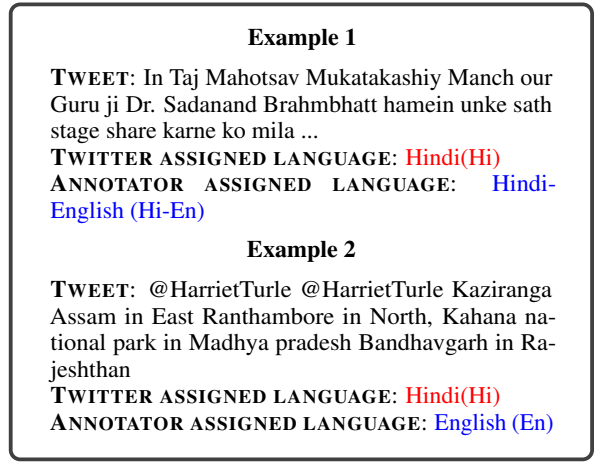

\centering
\small{
\begin{tcolorbox}[colback=white]
\begin{center}
    \textbf{Example 1}
\end{center}

\textsc{\textbf{Tweet}}: In Taj Mahotsav Mukatakashiy Manch our Guru ji Dr. Sadanand Brahmbhatt hamein unke sath stage share karne ko mila ...

\textsc{\textbf{Twitter assigned language}}: \textcolor{red}{Hindi(Hi)}

\textsc{\textbf{Annotator assigned language}}: \textcolor{blue}{Hindi-English (Hi-En)}

\begin{center}
    \textbf{Example 2}
\end{center}

\textsc{\textbf{Tweet}}: @HarrietTurle @HarrietTurle Kaziranga  Assam in East Ranthambore in North,  Kahana national park in Madhya pradesh Bandhavgarh in Rajeshthan \\
\textsc{\textbf{Twitter assigned language}}: \textcolor{red}{Hindi(Hi)}

\textsc{\textbf{Annotator assigned language}}: \textcolor{blue}{English (En)}




\end{tcolorbox}}
\caption{Tweets from the \textit{MMT-LID} dataset with language tags from Twitter and the human annotator.}
\label{fig:tweet_example}
\end{figure}

\begin{table*}[!h]
\centering
\resizebox{\hsize}{!}{
    \begin{tabular}{l|l|c|c}
    \hline
    \textbf{Topic} & \textbf{Subtopics} & \textbf{\# Tweets} & \textbf{Avg len}\\ \hline
        Environment & Pollution, Climate Change, Eco Friendly, Floods & 142208 &216.58 \\ \hline
        
        Food & Online food delivery, Food security, Indian desserts& 195086 & 140.75 \\ \hline
        
        Economics and Retail &  \begin{tabular}[l]{@{}l@{}}Initial Public Offering (IPO), SEBI and New margin rules,\\ Unicorns, Unemployment in India \end{tabular} & 158016 & 179.16 \\ \hline
        
        Natural Disaster & Cyclone:, Earthquake, Pandemic, Flood & 75591 &161.18 \\ \hline
        
        Art and Literature & Forms of Indian Art, Art festivals, Literature festivals, Book Fairs & 111909 & 150.43 \\ \hline
        
        Sports & \begin{tabular}[l]{@{}l@{}}Olympics, Indian Premier League (IPL), Indian Super League (ISL), \\Pro Kabaddi League (PKL)\end{tabular} & 122740 & 117.06  \\ \hline
        
        Politics & Pegasus Snooping, Farmer Agitation, West Bengal Elections, 2021 & 119963 &155.0\\ \hline
        
        R\&D and Technology & Mobile Technology, Health-Tech and Medical Innovations, ISRO & 111615 & 166.43 \\ \hline
        
        Wildlife and Vegetation & \begin{tabular}[l]{@{}l@{}}Kaziranga National Park, Bandhavgarh National Park, Nilgiri National Park,\\ Corbett National Park, Ranthambore National Park, Gir National Park, Nanda\\ Devi National Park, Save Tiger Project, Save Elephants, Save the Great Indian\\ Bustard,  Wildlife Tourism and Heritage, Forest Cover, River Rejuvenation, \\Restoration, Wildlife Crime, Climate Change \end{tabular} & 280091 & 155.03 \\ \hline
        
        Manufacturing & \begin{tabular}[l]{@{}l@{}}Make in India, Steel Manufacturing, Automobile Manufacturing, \\Electronics and electrical manufacturing\end{tabular} & 100969 & 125.14\\ \hline

        Films and OTT & \begin{tabular}[l]{@{}l@{}}OTT platforms such as Netflix, Amazon Prime Video,  OTT Censorship, \\OTT Voicecalling, Nepotism in Film Industry, NationalFilmAwards \end{tabular} & 89760 & 142.12
        \\ \hline
        
        Journalism \& Media &  \begin{tabular}[l]{@{}l@{}}Policy and Trends, Print Media \& TV, Criminal Journalism,\\ Social Movements and News \end{tabular} & 139563 & 170.88 \\ \hline
        
        Education & Exams, IIT, Online Education, Education System & 107634 & 186.39 \\ \hline
    \end{tabular}}
\caption{Distribution of topics, subtopics, the number of tweets, and the average length of tweets in the \textit{MMT} dataset. By incentivizing teams to collect over 0.1 million, we obtained more than 0.1 million tweets for 11 seed topics.}
\label{tab: topics_dist}
\end{table*}

\subsection{MMT-LID} 
We construct this dataset using a language annotation task on the MMT dataset.  We assign each team member (of the 13 teams) a randomly selected set of 500 tweets (with no duplicates) from the same seed topic as assigned in the \textit{MMT's data collection} step. We provide the following guidelines for the annotation task:

\begin{itemize}[noitemsep,nolistsep,leftmargin=*]
\item For each selected tweet, mark if the Twitter-assigned language tag is correct. In case the tag is incorrect, identify the  correct language tag. In case the text mixes multiple languages, assign a combined tag by separating them using a hyphen. For example, if the tweet text mixes Hindi (either in Devanagari or Roman) and English tokens, the first answer will be `\textit{No}’, and the second answer will be `\textit{Hi-En}’.
\item In case the tweets are code-mixed,  identify and annotate the main language (whose grammar is followed) and the embedded language (whose few tokens are embedded in the main language). For example, in the tweet ``\textit{items ko cart me daal ke app band kar dena is not funny}'', the main language is ‘\textit{Hi}’ and embedded language is ‘\textit{En}’.
\end{itemize}

As a result of the annotation, we obtain 5,346 tweets with human-annotated language tags. To evaluate the annotator's performance on this task, we evaluate the inter-annotator agreement (IAA) for each of the 13 topics using Cohen's Kappa (CK) score. We re-annotate 325 tweets (25 randomly selected tweets from each topic of the MMT-LID dataset) with the language tags and then calculate CK for IAA. Overall, we achieve an IAA score of 0.94. In Table \ref{tab: top_5_lang}, we report IAA scores per topic.

\begin{table*}[!t]
\centering
\resizebox{0.87\hsize}{!}{
\begin{tabular}{l|ccccccccc}
\hline
\textbf{Topic}                                                 & \textbf{English} & \textbf{Hindi} & \textbf{Bengali} & \textbf{Marathi} & \textbf{Telugu} & \textbf{Unidentified} & \textbf{\#L}  & \textbf{Avg len} & \textbf{IAA} \\ \hline
Environment   & 136929 &575& 5 & 22 & 8 & 2667       &45 & 216.58 & 0.96\\ \hline
Food    & 135094 &17141&125 & 366 & 96 & 10731 &45 &140.75  &0.91\\ \hline

\begin{tabular}[c]{@{}l@{}}Economics \& Retail \end{tabular}  &  141766 &4295&	19 & 71 & 15 & 4801   &45 & 179.16 & 0.94\\ \hline
Natural Disaster       &  37081  &16670& 740 & 547 & 1257 & 2915    &43  & 161.18 & 0.91\\ \hline
\begin{tabular}[c]{@{}l@{}}Art and Literature \end{tabular}&  85389 &2955& 139 & 63 & 79 & 6977 &44  &150.43 &0.93\\ \hline

Sports  &  56952  &5493& 519 & 129 & 83 & 9996  &45  & 117.06 & 0.94\\ \hline

Politics &  56469  & 25112&666 & 537 & 99 & 20532  &43 & 	155.0 & 0.89 \\ \hline

\begin{tabular}[c]{@{}l@{}}R\&D and   Technology \end{tabular} &  75176  &5428& 99 & 195 & 129  & 3377  &45 & 166.43 & 0.93\\ \hline

\begin{tabular}[c]{@{}l@{}}Wildlife \&  Vegetation \end{tabular} &  203024 &18831& 	42 & 591 & 24 & 7063   &45 & 155.03 & 0.90 \\ \hline
Manufacturing &  48421 & 1805&19 & 50 & 221 & 3220   &47 & 125.14 & 0.94\\ \hline
Films \& OTT          &  70314 &2808& 7 & 30 & 49 & 6274   &44 &142.12&  0.94\\ \hline
\begin{tabular}[c]{@{}l@{}}Journalism \& Media \end{tabular} &  80762  & 29663&838 & 1725 &  481 & 12703    &46 & 170.88 & 0.90\\ \hline
Education  &  80848  &7937&50 & 216 & 147 & 3435    &45 &186.39& 0.92 \\ \hline
\end{tabular}}
\caption{Topic-wise distribution of top-5 most spoken Indian languages (according to 2011 Census of India)
. \#L: number of unique languages, and Avg len: average length of tweets. 
}
\label{tab: top_5_lang}
\end{table*}

\subsection{Dataset Analysis}
\begin{figure}[!tbh]
  \centering
  \includegraphics[width=0.40\textwidth]{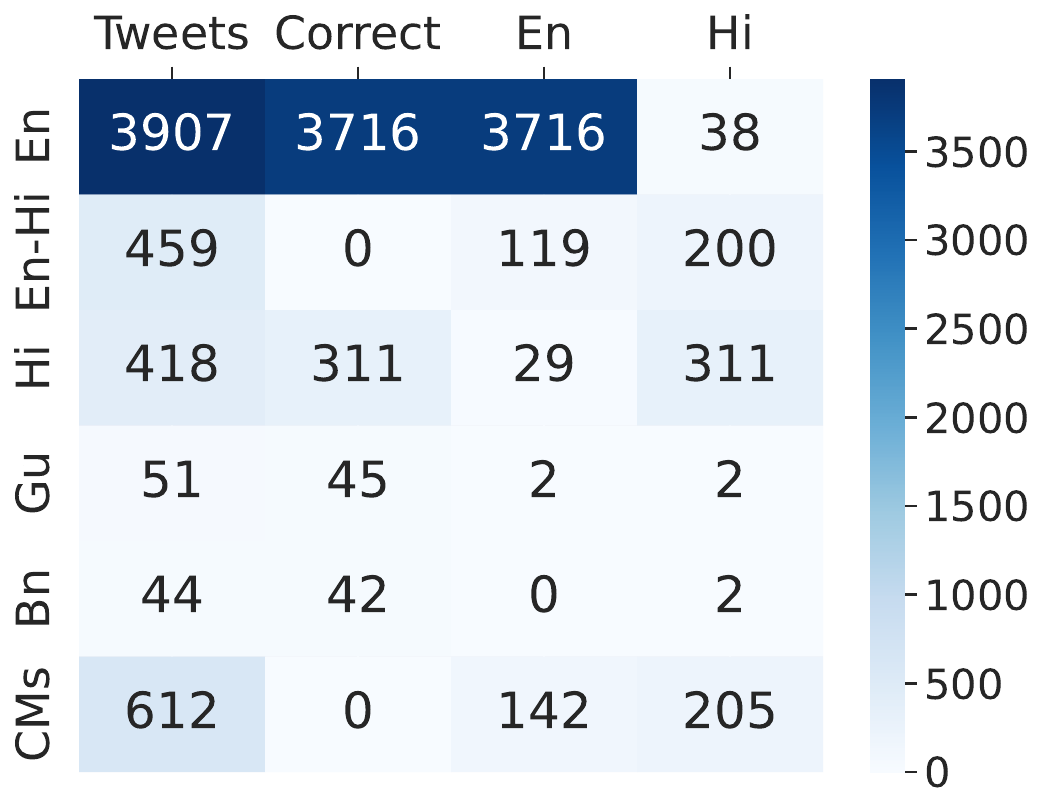}
  \caption{Distribution of language annotation by human annotators in the \textit{MMT-LID} dataset. Here, we report the top-5 identified languages by the human annotators in the \textit{MMT-LID} dataset. Here, \texttt{Correct} shows the number of tweets with correct language identification by Twitter. The column name \texttt{En} and \texttt{Hi} show the language identified by Twitter. \texttt{CMs} show all tweets in code-mixed languages.}
  \label{tab: lang_data}
\end{figure}

We make several interesting observations from the \textit{\textbf{MMT}} and \textit{\textbf{MMT-LID}} datasets. We list these observations below:

\begin{itemize}[noitemsep,nolistsep,leftmargin=*]
    \item Table~\ref{tab: top_5_lang} showcases that tweets for topics such as \textit{`Environment'}, \textit{`Education'},  and \textit{`Economics \& Retail'} are significantly longer than topics such as \textit{`Sports'}, \textit{`Manufacturing'}, and \textit{`Food'}. The significant difference in the average lengths illustrates the diversity in the discussions; for example, agendas, news, and political topics represent lengthier conversations than  match updates, movie reviews, and product launches.
    

    
    
    
    \item Figure \ref{tab: lang_data} shows the distribution of top-5 languages (as identified by the human annotators) in the \textit{\textbf{MMT-LID}} dataset. We observe that the majority ($\approx$95\%) of the English language tweets are correctly identified by Twitter. We identify that code-mixed language Hinglish is the second most frequent language in the dataset. TLID identifies the majority of the Hinglish tweets as either English or Hindi. We observe that 11.45\% of tweets in \textit{\textbf{MMT-LID}} dataset are code-mixed. This also includes tweets that mix English with other (non-Hindi) languages. Interestingly, we found 175 annotated tweets where none of the languages in the code-mixed pair were identified by TLID.  
\end{itemize}

\section{Answering the Pertinent Questions}
\label{sec:qna}
In this section, we explore the three research questions posed in Section \ref{sec: intro}.

\subsection{\textit{RQ1}: how  do traditional topic modeling tools perform in multilingual settings?}
\label{sec:traditional_TM}

We answer this question by exploring the traditional topic modeling algorithm LDA~\cite{blei2003latent}.
We conduct experiments on MMT and MMT-LID datasets based on the coarse and fine-grained topic categorization. For each experiment, we randomly partition the dataset into a 95:5 ratio, wherein a 95\% split is used for training the LDA model and 5\% for inference. We report the model's accuracy, weighted F1-score (W-F1), and coherence score~\cite{roder2015exploring} for each experiment.

\subsubsection{Inferring topics in MMT dataset}
\label{sec:coarse_MMT}
In the first experiment, we separately train the LDA model on the MMT dataset's train split with 13 topics and 63 subtopics. Each of the trained topics (and subtopics) is manually
assigned to one of the 13 original topics (and 63 subtopics). In Table~\ref{tab: result_topic}, we report the result of our experiment with the LDA topic model on the inference split of the MMT dataset.

In the second experiment, we partition the MMT dataset into two partitions based on language tags assigned by Twitter's language identification tool. The first partition comprises English tweets (1,208,225 tweets), and another partition comprises non-English tweets (546,920 tweets). For each partition, we follow the same steps as the first experiment (described above). The scores (see Table~\ref{tab: result_topic}) for the English partition are better than the non-English partition. We witness a significant drop in the accuracy and coherence scores in the non-English partition. This showcases the inefficacy of LDA in handling multilingual datasets. As English tweets are present in majority in the MMT dataset, we attribute this imbalance for higher scores of English against the full MMT dataset. 

\begin{table}[!t]
\centering
\resizebox{\hsize}{!}{
\small{
\begin{tabular}{c|c|cc|cc}
\hline
\multirow{2}{*}{\textbf{Language}}    & \multirow{2}{*}{\textbf{Metric}} & \multicolumn{2}{c|}{\textbf{13 topics}} & \multicolumn{2}{c}{\textbf{63 subtopics}} \\  \cline{3-6} 
&            &\multicolumn{1}{c|}{LDA}      & CTM     & \multicolumn{1}{c|}{LDA}       & CTM       \\ \hline
\multirow{3}{*}{\textbf{All}}         & Accuracy                                                                               & \multicolumn{1}{c|}{0.424}         & 0.492        & \multicolumn{1}{c|}{0.095}          & 0.130          \\  
& W-F1   & \multicolumn{1}{c|}{0.408}         & 0.469        & \multicolumn{1}{c|}{ 0.091}          & 0.124          \\    & Coherence             & \multicolumn{1}{c|}{0.534}         &   0.629      & \multicolumn{1}{c|}{0.542}          &   0.636        \\ \hline
\multirow{3}{*}{\textbf{En}}     & Accuracy                                                                               & \multicolumn{1}{c|}{0.443}         &  0.521       & \multicolumn{1}{c|}{0.102}          &  0.144         \\  
& W-F1    &  \multicolumn{1}{c|}{0.399}         &  0.478       & \multicolumn{1}{c|}{0.089}          &    0.128       \\ 
& Coherence       & \multicolumn{1}{c|}{0.573}         &  0.654       & \multicolumn{1}{c|}{0.590}          &  0.659         \\ \hline
\multirow{3}{*}{\textbf{Non-En}} & Accuracy                                                                               & \multicolumn{1}{c|}{0.398}         &      0.461   & \multicolumn{1}{c|}{0.084}          &    0.119       \\ 
& W-F1 & \multicolumn{1}{c|}{0.379}         & 0.437        & \multicolumn{1}{c|}{0.086}          &    0.113       \\ 
& Coherence   & \multicolumn{1}{c|}{0.384}         &  0.512       & \multicolumn{1}{c|}{0.407}          &  0.563         \\ \hline
\end{tabular}}}
\caption{Perfomance evaluation of the topic modeling systems on the \textit{MMT} dataset.}
\label{tab: result_topic}
\end{table}

\subsubsection{Inferring topics in MMT-LID dataset}
\label{sec:coarse_MMT_LID}
Next, we conducted two similar experiments (described in the previous section) on the MMT-LID dataset. The main motivation for conducting these experiments is to bypass the errors introduced by Twitter's language identification tool. The results (see Table~\ref{tab: result_topic_lid}) follow the experimental observations conducted in the previous section. In comparison to non-English multilingual datasets, LDA performs better on monolingual English datasets. We believe that the small size of the dataset led to the discrepancy in the coherence score. The small size dataset limits the number of words for the model to learn. Thereby limiting the number of coherent words in a topic cluster, making the coherence score very volatile and dataset dependent~\cite{syed2017full}.

\subsection{\textit{RQ2}: can we achieve better topic modeling with the cross-lingual contextual topic model (CTM)?}
\label{sec:contextual_TM}
The pertinent problem in the traditional LDA model lies with the bag-of-word (BoW) assumption, which disregards grammar and word order and only considers  the frequency of words. As a result, such topic models cannot effectively deal with unseen words in the document. Additionally, such topic models do not perform well on  multilingual corpora without combining the vocabulary of multiple languages. To overcome these challenges, we experiment with \textbf{ZeroShotTM}~\cite{bianchi2021cross}, which is a cross-lingual contextual topic model supporting multilingual embeddings. 

\begin{table}[!t]
\centering
\resizebox{\hsize}{!}{
\small{
\begin{tabular}{c|c|cc|cc}
\hline
\multirow{2}{*}{\textbf{Language}}    & \multirow{2}{*}{\textbf{Metric}} & \multicolumn{2}{c|}{\textbf{13 topics}} & \multicolumn{2}{c}{\textbf{63 subtopics}} \\  \cline{3-6}
                                      &                                                                                        & \multicolumn{1}{c|}{LDA}      & CTM     & \multicolumn{1}{c|}{LDA}       & CTM       \\ \hline
\multirow{3}{*}{\textbf{All}}         & Accuracy                                                                               & \multicolumn{1}{c|}{0.395}         &      0.488   & \multicolumn{1}{c|}{0.090}          &       0.141    \\ 
                                      & W-F1                                                                             & \multicolumn{1}{c|}{0.363}         &   0.434     &  \multicolumn{1}{c|}{0.082}          & 0.129          \\  
                                      & Coherence                                                                              & \multicolumn{1}{c|}{0.447}         &      0.602   &  \multicolumn{1}{c|}{0.442}          &   0.619        \\ \hline
\multirow{3}{*}{\textbf{En}}     & Accuracy                                                                               & \multicolumn{1}{c|}{0.472}         &   0.637      & \multicolumn{1}{c|}{0.139}          &      0.193     \\  
                                      & W-F1                                                                            & \multicolumn{1}{c|}{0.448}         &     0.591    & \multicolumn{1}{c|}{0.126}          &        0.179   \\  
                                      & Coherence                                                                              & \multicolumn{1}{c|}{0.386}         &    0.589    & \multicolumn{1}{c|}{0.418}          & 0.624           \\ \hline
\multirow{3}{*}{\textbf{Non-En}} & Accuracy                                                                               & \multicolumn{1}{c|}{0.297}         &   0.442      & \multicolumn{1}{c|}{0.061}      &  0.110          \\ 
                                      & W-F1                                                                           & \multicolumn{1}{c|}{0.301}         &    0.381     & \multicolumn{1}{c|}{0.064}          & 0.102        \\  
                                      & Coherence                                                                              & \multicolumn{1}{c|}{0.546}         &    0.667      & \multicolumn{1}{c|}{0.553}          & 0.676           \\ \hline
\end{tabular}}}
\caption{Perfomance evaluation of the topic modeling systems on the \textit{MMT-LID} dataset.}
\label{tab: result_topic_lid}
\end{table}

We conduct similar experiments described in Section~\ref{sec:traditional_TM} by replacing traditional LDA with ZeroShotTM. Tables~\ref{tab: result_topic} and \ref{tab: result_topic_lid} showcase the higher of ZeroShotTM (labeled as CTM) against LDA. However, the performance under the multilingual non-English partition is still significantly lower than the monolingual English partition.

\subsection{\textit{RQ3}: how do multilingual language identification tools perform in the multi-topical text?}
\label{sec:SOTA_LID}
Here, we explore the performance of the multilingual language identification systems on \textit{MMT-LID} dataset. We experiment with four language identification systems as given in \cite{srivastava2021challenges}, i.e., 
Polyglot,
FastText,
Langdetect,
and CLD3.
\begin{table}[!b]
\centering
\resizebox{\hsize}{!}{
\small{
\begin{tabular}{c|c|c|c|c|c|c}
\hline
\textbf{Language}   &  \textbf{Metric} & \textbf{TW} & \textbf{PG} & \textbf{FT} & \textbf{LD} & \textbf{CLD3} \\ \hline
\multirow{2}{*}{\textbf{All}}         & Accuracy                                                              &0.816 & 0.812 & 0.820 & 0.797 & 0.721          \\  & W-F1   & 0.795 & 0.777 & 0.780 & 0.781 & 0.755           \\ \hline
\multirow{2}{*}{\textbf{En}}     & Accuracy               & 0.945 & 0.973 & 0.983 & 0.957 & 0.856     \\ 
& W-F1                            & 0.972 & 0.986 & 0.991 & 0.978 & 0.922             \\ \hline
\multirow{2}{*}{\textbf{Non-En}} & Accuracy                                                            & 0.462 & 0.372 & 0.379 & 0.360 & 0.352              \\  
& W-F1                           & 0.392 & 0.362 & 0.349 & 0.352 & 0.348           \\ \hline
\end{tabular}}}
\caption{TW: Twitter, PG: Polyglot, FT: FastText, LD: Langdetect and CLD3: Compact Language Detector v3.}
\label{tab: results_LID}
\end{table}

In addition, we report the performance of the TLID. We use the language tags assigned by the human annotators as a reference for evaluation. To report the system performance, we use two evaluation metrics, i.e., accuracy and weighted F1 score. Table \ref{tab: results_LID} shows the results of multilingual language identification systems on the \textit{MMT-LID} dataset. We observe that all the systems perform extremely well on the English dataset. We observe a drop in system performance with the entire \textit{MMT-LID} dataset. Also, with only non-English data, all the systems show extremely poor results. These results indicate that multilingual language identification tools perform poorly in real-world settings where data from multiple languages and topics co-exist.

\section{Limitations and Future Works}

We collected the dataset from Twitter without language-specific constraints to reflect the real-world distribution of languages. This means that English, as a primary language, is over-represented in the dataset, while under-spoken languages such as Assamese are under-represented due to their limited use on the platform. This difference in distribution presents a challenge for building a robust multilingual system that performs well for such underrepresented languages. To overcome this, data augmentation techniques such as paraphrasing and oversampling, as well as transfer learning methods, can be utilized. These techniques can help balance the representation of languages in the dataset and further improve the performance of the multilingual system.

\section{Concluding Remarks}
\label{sec:challenge}
In this paper, we present a multilingual and multi-topic dataset collected from Twitter for the Indian community spanning various Indian languages, including but not limited to the popular code-mixed languages. This could prove useful for further understanding and exploring the natural phenomenon of the co-existence of multilingual and multi-topical data. We also showcased several issues in topic modeling the multilingual dataset using traditional algorithms like LDA. We believe that the availability of such a large-scale and quality dataset will be useful in building systems for numerous downstream tasks such as multilingual topic modeling, language identification, machine translation, etc.

\bibliographystyle{ACM-Reference-Format}
\bibliography{sample-base}

\flushend
\end{document}